# Ensemble Classifier for Eye State Classification using EEG Signals

**Ali Al-Taei[1]**

IT Dept., Technical College of Management,

Middle Technical University, Baghdad, Iraq

**Abstract:** The growing importance and utilization of measuring brain waves (e.g. EEG signals of eye state) in brain computer interface (BCI) applications highlighted the need for suitable classification methods. In this paper, a comparison between three of well-known classification methods (i.e. support vector machine (SVM), hidden Markov map (HMM), and radial basis function (RBF)) for EEG based eye state classification was achieved. Furthermore, a suggested method that is based on ensemble model was tested. The suggested (ensemble system) method based on a voting algorithm with two kernels: random forest (RF) and Kstar classification methods. The performance was tested using three measurement parameters: accuracy, mean absolute error (MAE), and confusion matrix. Results showed that the proposed method outperforms the other tested methods. For instance, the suggested method's performance was 97.27% accuracy and 0.13 MAE.

## 1. Introduction

Eye state detection is the task of predicting the state of eye whether it is open or closed. To achieve this task, a new trend of using brain activity signals by the mean of electroencephalography (EEG) measures for the training and testing of various machine learning classification algorithms was investigated by many researchers [1, 2]. However, the task of predicting human actions via brain signals takes high importance and usability in various fields such as computer games [3-6], health care and bio-medical systems [7-9], emotion tracking [10], smart home device controlling and internet of things (IoT) [11, 12], military [13], and detection of car driving drowsiness [14]. To achieve this goal, employing a suitable classification algorithm is needed. However, although there are many classification methods, the performance of any method in a given task might not be sufficiently good in another task [15].

[1] muthqal@yahoo.com , alitaei@mtu.edu.iq

### 1.1 Objectives
The main aims of this paper are to (*i*) evaluate the performance of a combination of classification algorithms in order to predict the state of human eye based on brain signals, and (*ii*) developing a new classification method that is more robust and higher in performance.

### 1.2 Organization
The rest of this paper will be organized as the following sections. In section 2, the dataset characteristics will be illustrated. In section 3, the previous related works will be explained. In section 4, the experiments and results will be presented and discussed. In section 5, the conclusion is presented, followed by ideas for future work.

## 2. Dataset Characteristics
An Emotiv headset device with 16 sensors have been used to record brain signals. The duration time of each recording was 117 seconds. Then, the different eye states observed during each recording were manually added. Finally, the corpus dataset was prepared which contains 14980 instances. Each instance consists of 14 EEG features and an eye-state class (either 0 for open, or 1 for closed). Furthermore, the number of instances with open-eye class in the corpus dataset is 8257 (55.12%), while the number of closed-eye type instances is 6723 (44.88%). The dataset was created by Rösler and Suendermann, which was firstly used by them in [1]. Additionally, this dataset was also used in many researches such as [2, 16, 17]. However, according to [1] three of the instances' (2 open states, and one closed state) values were outliers so it is preferable to delete them before the classification process. In this research, the whole data set has been used in order to (*i*) taking in place the noise, and (*ii*) dealing with noisy data in the classification process.

## 3. Related work
There are many previous studies that focused on the detection of eye state from EEG signals. [17] proposed a neural fuzzy approach to predicting eye state from EEG signals. The error rate of this approach was 4%. [18] found that the power of the closed eye is higher than that of the open eye state. [19] used EEG based eye state dataset to track and detect eye blinking. In order to achieve this goal, they employed artificial neural network (ANN) approach. However, the results were very poor. In [1] a dataset of 14977 instances was collected and used to evaluate the performance of 42 classification methods to detect eye state. Results showed that Kstar is the best method for this task with an error rate of 2.7%. [20] compared between the performance of Instance Based (IBK) classifier and other trusted classifiers. Results showed that Random Forest (RF) and IBK performed higher than other classifiers. For instance, the highest performance obtained is for IB classifier with 94.5%. [21] tried Kstar classifier with an incremental feature reordering (IFR) method for the purpose of selecting/removing feature subset to eye state classification task. However, the performance of the suggested method was about 95%. [22] proposed a method based on incremental attribute learning (IAL) and neural network to predict eye state. Results showed that IAL was the lowest error rate classifier in comparison with other conventional classifiers. However, the performance of IAL depends on well-ordered features and proper training on those extracted features as well. [23] suggested using deep learning approach to achieve this goal. For instance, deep belief network (DBN) and stacked auto encoder (SAE) classifiers were experimented. Results showed that utilizing deep learning neural network approach to detecting the state of the eye is a promising method. However, the complexity and time factors have not been discussed.

## 4. Results and Discussion

In this section, three different classification algorithms will be employed and evaluated according to their ability of detecting eye state from EEG signals. For instance, the evaluated classifiers are support vector machine (SVM), hidden Markov map (HMM), and radial basis function (RBF). Furthermore, our proposed ensemble classifier will be explored and evaluated. The stratified cross validation method with 10 folds was used in all experiments. In addition, analysis and comparison between the employed classifiers' results will be illustrated by means of accuracy and mean absolute error (MAE).

### 4.1 Experimental Results

#### 4.1.1 Support Vector Machine (SVM)

The SVM classifier is tested and the output is presented in the confusion matrix form in Table 1.

Table 1: SVM confusion matrix

| Open | Close | |
|------|-------|-------|
| 8257 | 0 | Open |
| 6723 | 0 | Close |

From the result listed in Table 1 above, it is obvious that 8257 out of 14980 instances were classified correctly.

### 4.1.2 Hidden Markov Map (HMM)

The HMM classifier is performed on the dataset and the result is shown in Table 2 in the form of confusion matrix.

Table 2: HMM confusion matrix

| Open | Close | |
|------|-------|-------|
| 8257 | 0 | Open |
| 6723 | 0 | Close |

As shown in Table 2, the correctly classified instances are 8257 out of 14980. However, it can be noticed that SVM and HMM algorithms behave similarly in this case. For instance, all instances were classified as 'open' class.

### 4.1.3 Radial Basis Function (RBF)

The confusion matrix shown in Table 3 illustrates the output of RBF classifier.

Table 3: RBF confusion matrix

| Open | Close | |
|------|-------|-------|
| 5939 | 2318 | Open |
| 3993 | 2730 | Close |

From Table 3, it can be observed that 8889 out of 14980 instances were classified correctly. Despite this result, statistically, better than the results of the other previous classifiers, its performance still lower than other approaches in literature.

### 4.1.4 Proposed method (Vote (Kstar and RF))

The suggested method based on employing ensemble classifier (i.e. using multi kernel method) for the detection of eye state. In this paper, voting algorithm was employed with two kernels: Kstar and RF. In order to achieve this goal with highest performance, we trained/tested our kernel classifiers with different architectures and parameter values. However, results showed that the best classifier that performs the highest and less time consuming is based on Kstar and RF methods. For best results, here we employ RF with 180 iterations. Table 4 illustrates the confusion matrix of the proposed method.

Table 4: Proposed method

| Open | Close | |
|------|-------|-------|
| 8102 | 155   | Open  |
| 254  | 6469  | Close |

The results shown in Table 4 illustrate that 14561 out 14980 instances were classified correctly.

## 4.2 Discussion

According to the results obtained from the experiments in the previous subsection **4.1**, the classification accuracy and error rate (i.e. MAE) of each classifier were calculated and compared. Table 5 below, illustrates these results.

Table 5: Classification accuracy and MAE

| Classifier      | Accuracy (%) | MAE  |
|-----------------|--------------|------|
| SVM             | 55.12        | 0.45 |
| HMM             | 55.12        | 0.5  |
| RBF             | 57.87        | 0.47 |
| Proposed method | 97.27        | 0.13 |

From Table 5, it is obvious that the proposed method outperformed the other tested classifiers. For instance, the proposed method of voting Kstar and RF (with 180 iterations) performance is 97.27%. This result outperforms the both results of classifiers (i.e. Kstar performance is 96.77%, and RF performance is 93.84%) when employed independently. Furthermore, this result outperforms the results obtained in other works such as [1, 17, 20, 21, 24, 25]. However, although [1] suggested to use Kstar (with b=40) and obtained performance with less than 3% error rate, but the classifier was more complex and time consuming. In this paper, the proposed method's performance was slightly lower, which is due to using the corpus dataset with (3) noisy instances (as mentioned in Section 2). This result highlighted the possibility of working on (a little noisy) EEG eye state data to be detected accurately. Also, the proposed method is less complex (as we used Kstar with b=20).

## 5. Conclusion

This paper demonstrates that the results of the ensemble classification system are promising for the task of EEG based human eye state detection. For instance, a combination of two classification algorithms was applied (i.e. Kstar and RF) and the experimental result show that employing ensemble (multi-core) classifiers for the aim of eye state detection is better than the ordinary approach of using a single classification method. In addition, the proposed method's performance is higher than other previous methods (such as [17, 20-22, 26]). Furthermore, the proposed method's results highlight and encourage the area of real time eye state detection and the ability to work with incomplete/missing data.

Accordingly, ideas for future work we recommend collecting such datasets by recruiting more subjects, different EEG devices, and trying different sensors' positions. In addition, it is important to work on increasing the performance of the employed classifier and at the same time decreasing the processing time.